\documentclass[smallextended]{svjour3}       
\smartqed  

\usepackage{amsmath}
\usepackage[colorlinks]{hyperref}
\usepackage{graphicx}
\usepackage{graphics}
\usepackage{verbatim}
\usepackage{float}
\usepackage{mathtools}
\usepackage{mathptmx}
\usepackage{fullpage}

\usepackage[utf8]{inputenc} 
\usepackage[T1]{fontenc}    
\usepackage{hyperref}       
\usepackage{url}            
\usepackage{booktabs}       
\usepackage{amsfonts}       
\usepackage{nicefrac}       
\usepackage{microtype}      
\usepackage{xcolor} 
\usepackage{multirow}        
\usepackage{subfigure}
\usepackage{graphics}
\usepackage{graphicx}
\usepackage{amsmath}

\journalname{}

\makeatletter
\g@addto@macro{\UrlBreaks}{\UrlOrds}
\makeatother

\begin{document}

\title{Integrated Brier Score based Survival Cobra - A regression based approach} 

\titlerunning{IbsSurvCoBRA}        
\author{ Rahul Goswami and Arabin Kumar Dey
}

\institute{Rahul Goswami \at
             Department of Mathematics,\\ 
             IIT Guwahati,\\   
             Guwahati, India\\
             \email{arabin@iitg.ac.in} \\
             \and
             Arabin Kumar Dey \at
             Department of Mathematics,\\ 
             IIT Guwahati,\\   
             Guwahati, India\\
             \email{arabin@iitg.ac.in} \\
}

\maketitle

\begin{abstract}
  
  Recently Goswami et al. \cite{goswami2022concordance} introduced two novel implementations of combined regression strategy to find the conditional survival function.   The paper uses regression-based weak learners and provides an alternative version of the combined regression strategy (COBRA) ensemble using the Integrated Brier Score to predict conditional survival function.  We create a novel predictor based on a weighted version of all machine predictions taking weights as a specific function of normalized Integrated Brier Score.  We use two different norms (Frobenius and Sup norm) to extract the proximity points in the algorithm.  Our implementations consider right-censored data too.   We illustrate the proposed algorithms through some real-life data analysis.
  
\keywords{Survival Tree; COBRA; Integrated Brier Score; Right Censored Data}
\end{abstract}

\section{Introduction}
\label{intro}

 Ensembled techniques (\cite{dietterich2000ensemble}, \cite{giraud2014introduction}, \cite{shalev2014understanding}) are popular in estimating Conditional survival function.  Researchers (\cite{bellot2018boosted}, \cite{solomatine2004adaboost}, \cite{djebbari2008ensemble}) used different variations of Adaboost in their research work.  Boosting techniques have other several variants used in the same context, e.g. XGboost (\cite{chen2015xgboost}, \cite{mojirsheibani1999combining}).
 In this paper we formulate our solution within combined regression framework (COBRA) using Integrated Brier Score (IBS), which is not available in the literature.  
 
   The problem of predicting conditional survival function is different than predicting a response in general regression set up.  Since, we plan to predict a function of the response variable instead of a single response variable, our output space becomes high-dimensional or a vector of points,  which represents a function.  COBRA uses final prediction based on the collective response variables available in the proximity of the query point.  Finding this proximity set for each query point requires a metric to evaluate the similarity between two sets of predictors, which would be a matrix in this case.  Thus, we propose to use two matrix norms (Frobenius and Supremum norm) to carry out the algorithms.  
 
The problem in usual conditional survival estimation through combined regression strategy is it may not find any point in the proximity set.   We investigate such cases over different datasets and find ways to rectify the problem.  With a high probability there would be cases where only a few observations may be available in the proximity set, which can restrict efficient calculation of the final conditional survival function through Kaplan Meier or other similar methods.  We propose a new weighted version of all predictions from weak learners based on a specific function of IBS.  In this process, we do not need to calculate the survival based on the few observations obtained in the proximity set.   We observe that even if there is one observation in the proximity set, proposed weighted method can efficiently calculate the conditional survival function without hurting the notion of original COBRA. 
 
 We organize the paper in the following way.  In section 2, we provide the proposed Survival COBRA.  An illustration of the real dataset is available in section 3.  We conclude the paper in section 4.

\section{Proposed Survival COBRA}  

\subsection{Original Cobra}

 Usual combined regression strategy calculates the predictor at a particular instance as weighted sum of those test-sample, where weight calculation use the number of test sample instances, which are in the neighbourhood of predictions made by each predictors at that particular instance.  Suppose, we have the following N independent observations : $(Z_{i}, y_{i})\in R^{d + 1}; i = 1, \ldots, N$.  We divide the data set into training and test set, where test set observations are, $(Z_{i}, y_{i})\in R^{d + 1}; i = 1, \ldots, l$.  Let's denote the M weak learners that COBRA uses are, say $r_{1}(\cdot), r_{2}(\cdot), r_{3}(\cdot), \ldots, r_{M}(\cdot) : R^{d} \rightarrow R.$  The following is the mathematical expression of the COBRA strategy that expresses a predictive estimator $T_{n}$ : 
$$T_{n}(r_{k}(\textbf{x})) = \sum_{i = 1}^{l} W_{n, i}(\textbf{x}) Y_{i}, ~~~~~~ x \in R^{d}$$  

  Note that the random weights $W_{n, i}(\textbf{z})$ take the form :
$$ W_{n, i}(\textbf{z}) = \frac{1_{\cap_{m = 1}^{M} { \mid r_{k, m}(z) - r_{k, m}(Z_{i}) \mid \leq \epsilon_{l}}}}{\sum_{j = 1}^{l} 1_{\cap_{m =1}^{M} { \mid r_{k, m}(z) - r_{k, m}(Z_{i}) \mid \leq \epsilon_{l}} }}.$$

  Here $\epsilon_{l}$ is a positive number and $\frac{0}{0} = 0$ (by convention).  In addition, we can provide an alternative way to express COBRA when all original observations are invited to have the same, equally valued opinion on the importance of the observation $X_{i}$ (within the range of $\epsilon_{l}$).  Generally, $T_{n}$ includes the corresponding $Y_{i}$s in such cases.  However, the unanimity constraint usually relaxed by imposing a fixed fraction $\alpha \in \{ \frac{1}{M}, \frac{2}{M}, \cdots, 1 \}$ of the machines which agrees on the importance of $X_{i}$.  The weights then take the following form :
$$ W_{n, i}(x) = \frac{1_{\sum_{m = 1}^{M} 1_{ \mid r_{k, m}(x) - r_{k, m}(X_{j}) \mid \leq \epsilon_{l} } \geq M\alpha}}{\sum_{j = 1}^{l} 1_{\sum_{m = 1}^{M} 1_{ \mid r_{k, m}(x) - r_{k, m}(X_{j}) \mid \leq \epsilon_{l} } \geq M\alpha}}  $$  


  Since algorithm may not always be capable of selecting a non-zero number of similar samples if the threshold value goes much closer to zero or exactly equals to zero.   Increasing the testset size can remove the problem.  However, using a proportion ($\alpha$) of machine predictions instead of considering all the machine predictions could be an alternative solution. 
 
  We propose two Survival COBRAs in this context.  The conditional survival prediction is different than usual regression as it predicts a function of the response instead of the response.  It makes the output variable/prediction space always higher dimensional (since a set of points represents the function).  This functional structure demands necessary changes in the usual COBRA structure too.

\subsection{Straight IBS Survival Cobra}

  We divide the whole set as the train and test set with a specific split division and start optimizing the parameters based on the train set.  The algorithm will further subdivide the whole dataset into two parts, (say, $D_{l}$ and $D_{k}$), where all the weak-learners training use $D_{k}$ only.  In this process, the test set would behave like a query set or validation set, and we calculate the prediction based on $D_{l}$.  Since a small dataset would be much smaller in size due to such division, it may not allow some time to get the desired optimal parameters.  Therefore, we choose a different split division scheme.  We propose to use the whole dataset to find the optimal parameters instead of just using the train set and then split the dataset into training and test sets to get the survival estimate at optimized COBRA parameters.  

  The proposed algorithm in this paper uses weak learners as different survival trees.  We use eight weaker-learners to show the results.   Since each weak-learners reproduce survival functions, the output dimension is high (depending on the number of time point considered).  While aggregating the similar observations in the test set, we first need to choose a suitable norm to measure the similarity.  We use the Frobenius norm and Sup norm to obtain the similarity.  In general, we can obtain the value of the threshold parameter ($\epsilon_{l}$) and the number of predictors ($\alpha$) by minimizing the cross-validated test error of the prediction.  Since our prediction is a function, we propose to use a cross-validated Integrated Brier score for right censored data as an error function. 

  In our setup, we use Nelson-Allen estimators for the calculation of survival prediction based on the time-to-events corresponding to the similar observations as query points.  
 
\textbf{Nelsen-Allen Estimator : }   We consider the following definition of Nelson-allen estimator to make all cumulative hazard function calculation. 

$H(t \mid x^{i}) = \sum_{t_{i} < t}^{} \frac{d_{i}}{n_{i}}  $ where, $d_{i}$ is the number of deaths at time $t_{i}$ and $n_{i}$ is the number of people exposed to the risk of death at time $t_{i}$. 
    
\textbf{Integrated Brier Score }  We consider the benchmark as Integrated Brier Score ($IBS^{c}$) for censored observation which is slightly different than usual expression of Integrated Brier Score (IBS).  We can define Integrated Brier Score for censored observation between time interval $[t_{1} ; t_{\max}]$ as $ IBS^{c} = \frac{1}{t_{\max} - t_{1}} \int_{t_{1}}^{t_{\max}} BS^{c}(t) dt $.  We estimate the integration numerically via trapezoidal rule.  

 Expression of Brier Score for the censored observation ($BS^{c}$) is :
 
 $BS^{c}(t) = \frac{1}{N}\sum_{i = 1}^{N} \left[ I(y_{i} \leq t, \delta_{i} = 1) (\frac{(0 - \hat{S}(t \mid x^{i}))^{2}}{\hat{G}(y_{i})})  + I(y_{i} > t) \frac{(1 - \hat{S}(t \mid x^{i}))^{2}}{\hat{G}(t)} \right] $  
 
where, $\hat{S}(t \mid x^{i})$ is the predicted conditional survival function and $\frac{1}{\hat{G}(t)}$ is the inverse probability of the censoring weight.  We assume $C > t$, where C is the censoring point.

 Note that we choose the COBRA parameters by minimizing cross-validated Integrated Brier Score within right-censored set up. 
 
 According to the above proposed development, COBRA helps to find out the homogeneous group of patients.  Therefore, we assume that conditional survival functions for particular patient is same as that of the other similar patients.  The approach has inherent drawback of inability to calculate the curve when number of similar observations is very less (say one) or the cases when no observation is available for optimal choice of the parameter in a particular dataset.  
 
 We propose to use the following modification/variation of the algorithm in those cases.  We consider exploration of this paper dedicated only with IBS score in different angles.   
 
\subsection{A different variation - Weighted IBS Survival Cobra}

  Major bottleneck in above Straight IBS Survival COBRA is that the approach is incapable of evaluating conditional survival curve if insufficient number observations from test set come in proximity of a given patient observation.  We suggest an weighted version COBRA to rectify the problem of insufficient number of observations in the proximity of the query point.  The procedure is super efficient as it is capable of evaluating survival curve even if there is only one non-zero observation available in the proximity of the query point. 
  
  Key algorithmic steps of the weighted version COBRA is as follows :

\begin{enumerate}  

\item Find optimized parameters from the whole data set with a $k$-fold cross-validation with suitable choice of $k$.

\item Divide the data set into train and test.   We use test dataset as query point and further divide train dataset into two subdivisions where we use one such subdivision for training of the weak learners and make the prediction based on the other subdivision.

\item Collect all similar points (there should be at least one) in the proximity of the query point based on the chosen optimized parameters.  Let's assume the points are $X_{1}, X_{2}, \cdots, X_{m}$

\item Aggregate each machine prediction (weak learners) on the set of similar points (proximity set) by taking average of all prediction on that proximity set.  $r^{'}_{j} = \frac{1}{m} \sum_{i = 1}^{m} r_{j}(X_{i});  j = 1, \cdots, M $.  Note that $r^{'}_{j}$ is vector of values evaluated at finite number of time points representing the survival function predicted by j-th machine. In case of usual regression it is scale quantity.  We assume all similar points are homogeneneous in nature and thus possess same survival curve.

\item Final prediction is then calculated as weighted average of all machine predictions (weak learners) obtained in the previous step where weights are taken as normalized integrated brier score by each machine. $r_{F} = \frac{1}{M} \sum_{k = 1}^{M} W_{k} r^{'}_{k} $, $ W_{k} = \frac{IBS_{k}}{\sum_{k}^{} IBS_{k}}$, $ W^{'}_{k} = 1 - W_{k}$ where $IBS_{k}$ is the IBS score of the k-th machine.

\end{enumerate}

 The above approach does not directly use the response variables in the weight calculation.  Since it combines predictions of different machines in the proximity points based on weighted combination of machines, final prediction calculation is capable of generating survival curve or cumulative hazard function even when there is only one observation in the proximity set.

\textbf{Key issues/Limitations with Weighted Survival Cobra}  Inclusion of large number of machines increase the computational runtime of the algorithm.  The algorithm does not fail only when it ensures inclusion of at least one observation in the proximity of the query point.  However, such situations are unlikely as we can always tune/ slightly perturb our optimal parameters to include at least one point in an extreme situations.  We did not face such situation in anyone of our data analysis.  

\section{Data Analysis}

\subsection{Data Set Descriptions}   All datasets mentioned below are publicly available and extracted directly from python module \textit{Scikit-survival 0.18.0}.  We consider three datasets for our experiment :
\begin{itemize}
    \item Dataset I :  Worcester Heart Attack Study dataset \label{data2}, it consists of 500 samples and 14 features. A total number of 215 deaths of patients are available.
    \item Dataset II :  German Breast Cancer Study Group 2 dataset \label{data4} : The dataset has 686 samples and 8 features.
    The endpoint is recurrence free survival, which occurred for 299 patients (43.6\%).  There are two response variables :
    $cens$: Boolean indicating whether the endpoint has been reached or the event time is right censored.    
    $time$: Total length of follow-up.
    \item Dataset III : Veteran Lung Cancer Data \label{data1}, it has 137 samples and 6 features out of which number of deaths recorded are 128.
\end{itemize}
      
\subsection{Numerical Results}  

  The codes are available/accessible in the following public repository of Github link: \\
  \url{https://github.com/medsolu/DoctorInfo-COBRAibs}. All codes are run at IIT Guwahati computers with the following configurations : (i) Intel(R) Core(TM) i5-6200U CPU 2.30 GHz.  (ii) Ubuntu 20.04 LTS OS, and (iii) 12 GB Memory.
  

  A total of 12 figures from Figure-\ref{fig:ibs1} to Figure-\ref{fig:ibs12} show the conditional survival function for different datasets, norms and two different implementations of the proposed algorithm.  Initially, the experiment uses 50\% of the initial division of the whole dataset to form $D_{l}$ and $D_{k}$ and perform optimization of the parameters.  We choose 80\% and 20\% division ratios for the train and test set for final prediction.  We consider eight survival trees as weak learners in our setup.  The algorithm minimizes the five-fold cross-validated IBS score on the test set to find the optimal parameter values.  
 
  IBS score calculations for different datasets and different norms are available in Table-\ref{table:ibs}.   The notations COBRA-1 and COBRA-2 use the Frobenius norm and Sup-norm, respectively.  We randomly select the first consecutive weak learners and verify that the ensemble COBRAs provide lower IBS scores than all weak-learners.  Since the performance of Cobra-1 and Cobra-2 are the same for each dataset, one can choose to use any one of the above-mentioned norms in practice.  We observe weighted COBRA performs slightly better or almost equal to the Straight Survival COBRA across all datasets.
  
\begin{table*}[ht!]
\caption{Integrated Brier Score for different COBRA algorithms and different Datasets}
\label{table-concor}
\centering    
\begin{tabular*}{\textwidth}{@{\extracolsep{\fill}}crrrrrc} \hline \\
\toprule
& \multicolumn{2}{c}{ DATASET I} & \multicolumn{2}{c}{ DATASET II} & \multicolumn{2}{c}{ DATASET III} \\
\\\cmidrule(r){2-3}\cmidrule(lr){4-5}\cmidrule(l){6-7}
 &  mean &  & mean &  & mean &  \\
 & sd.  &       & sd.   &   & sd &  \\
\midrule
\multirow{2}{*}{Weighted IBS Survival COBRA 1 } & 0.191 &  & 0.190  & & 0.126  & \\ 
&0.02 && 0.02 && 0.03 & \\ [2ex]
\multirow{2}{*}{Weighted IBS Survival COBRA 2 } &  0.195 &  & 0.190 & & 0.124 & \\ 
&0.02 && 0.02 && 0.03 & \\ [2ex]
\multirow{2}{*}{Straight IBS Survival COBRA 1 } &  0.198 &  & 0.192 & & 0.125 & \\ 
& 0.03 && 0.02	 && 0.03   & \\ [2ex]
\multirow{2}{*}{Straight IBS Survival COBRA 2 } &  0.209 &  & 0.196  & & 0.127 & \\ 
& 0.04 && 0.02 && 0.03 & \\ [2ex]
\bottomrule
\end{tabular*}
\label{table:ibs}
\end{table*}     

\section{Conclusion}  The paper shows successful integration of tuning the parameters using integrated brier score in a censored setup.  We propose a novel weighted architecture other than straight COBRA implementation.  Our weighted architecture calculates the conditional survival function even when only one point appears in the proximity of the query point.  Our proposed weighted version works better or equally well compared to Straight IBS Survival COBRA across all norms.  We can use soft weights through an appropriate choice of kernels instead of the hard choice of weight function.  The work is in progress.   

\bibliographystyle{spmpsci}
\bibliography{surv}


\begin{figure}[ht!]
 \begin{center}
  \subfigure[$\xi_{1}$]{\includegraphics[width = 0.45\textwidth]{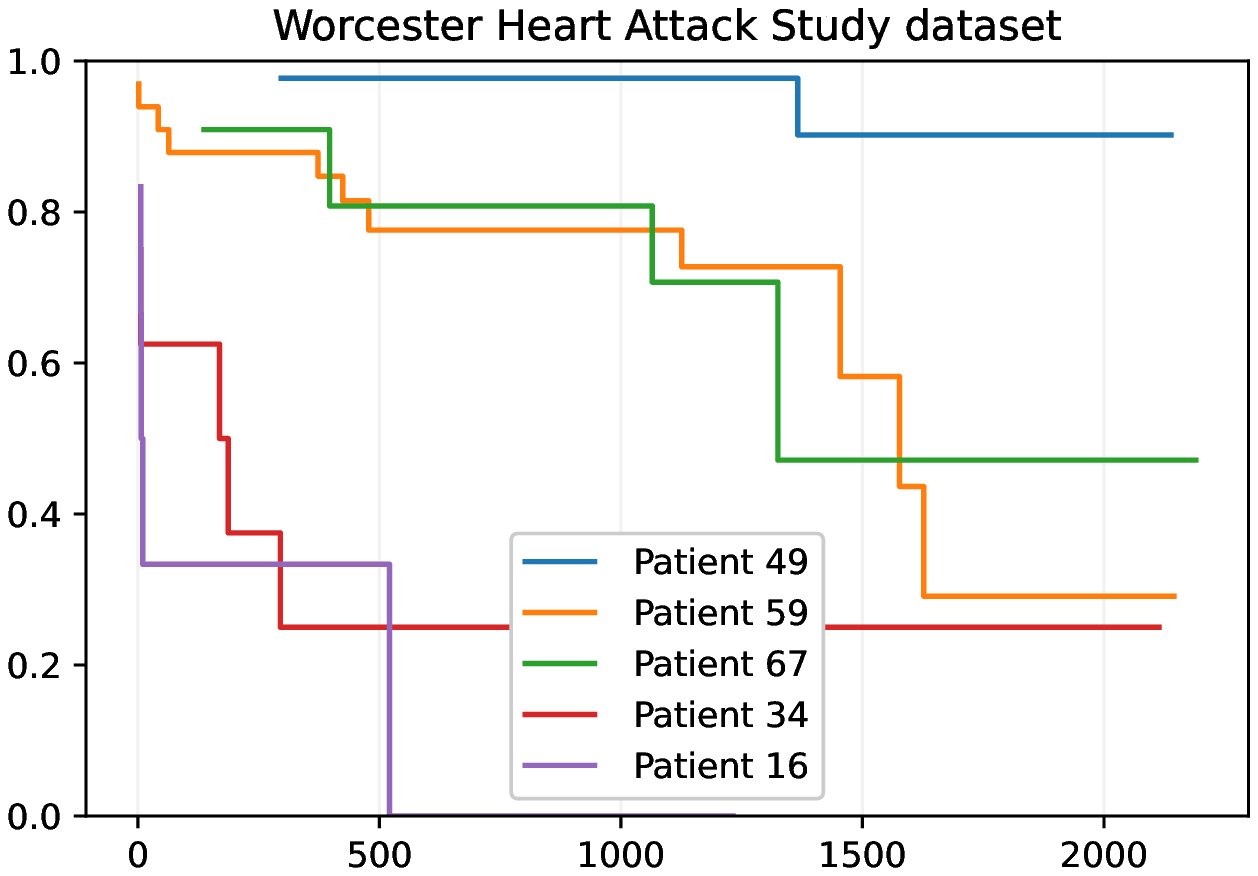}}
  \subfigure[$\xi_{2}$]{\includegraphics[width = 0.45\textwidth]{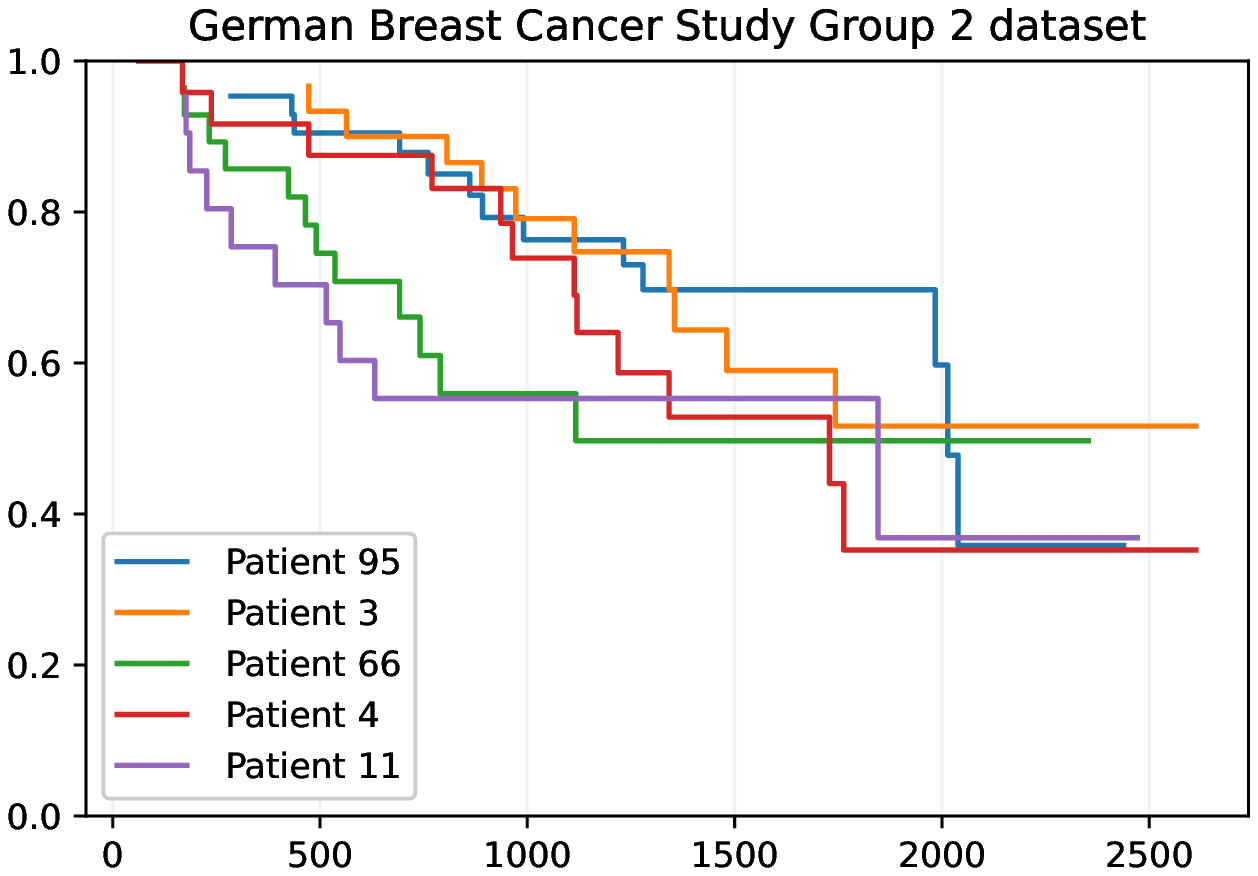}}\\
  \subfigure[$\xi_{3}$]{\includegraphics[width = 0.65\textwidth]{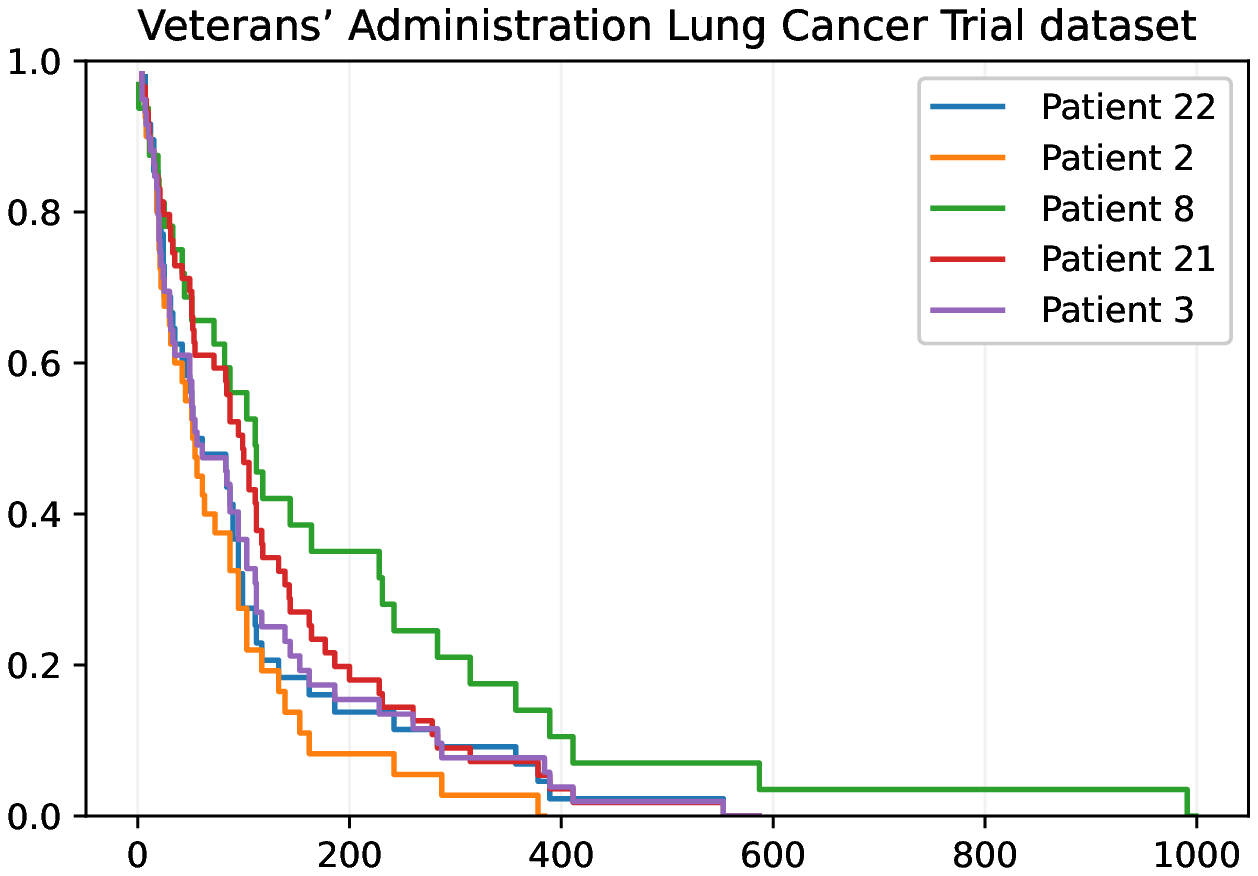}}\\
\caption{Survival Curve for Vanilla Survival Cobra for three datasets with Frobenius\_Norm \label{fig1}}
\end{center}
\end{figure}

\begin{figure}[ht!]
 \begin{center}
  \subfigure[$\xi_{1}$]{\includegraphics[width = 0.45\textwidth]{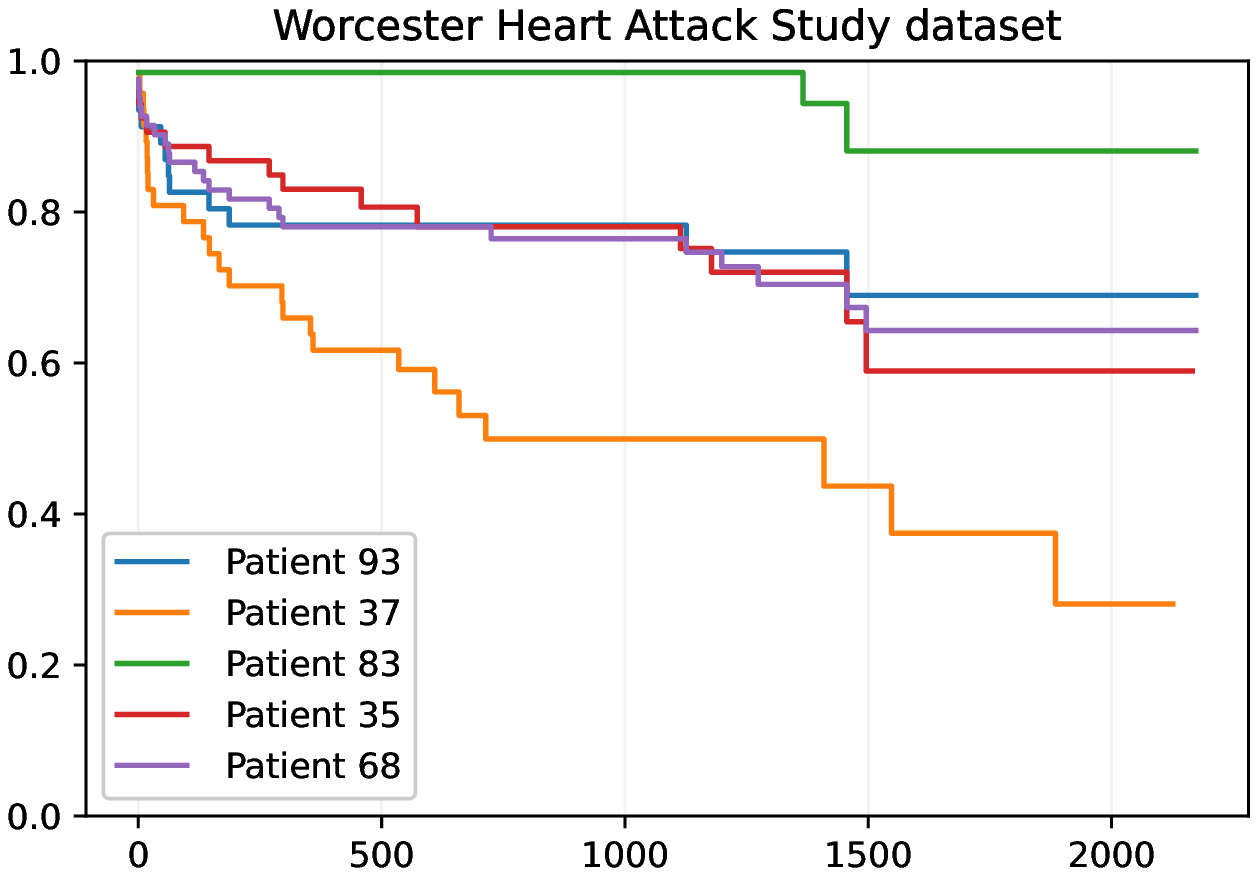}}
  \subfigure[$\xi_{2}$]{\includegraphics[width = 0.45\textwidth]{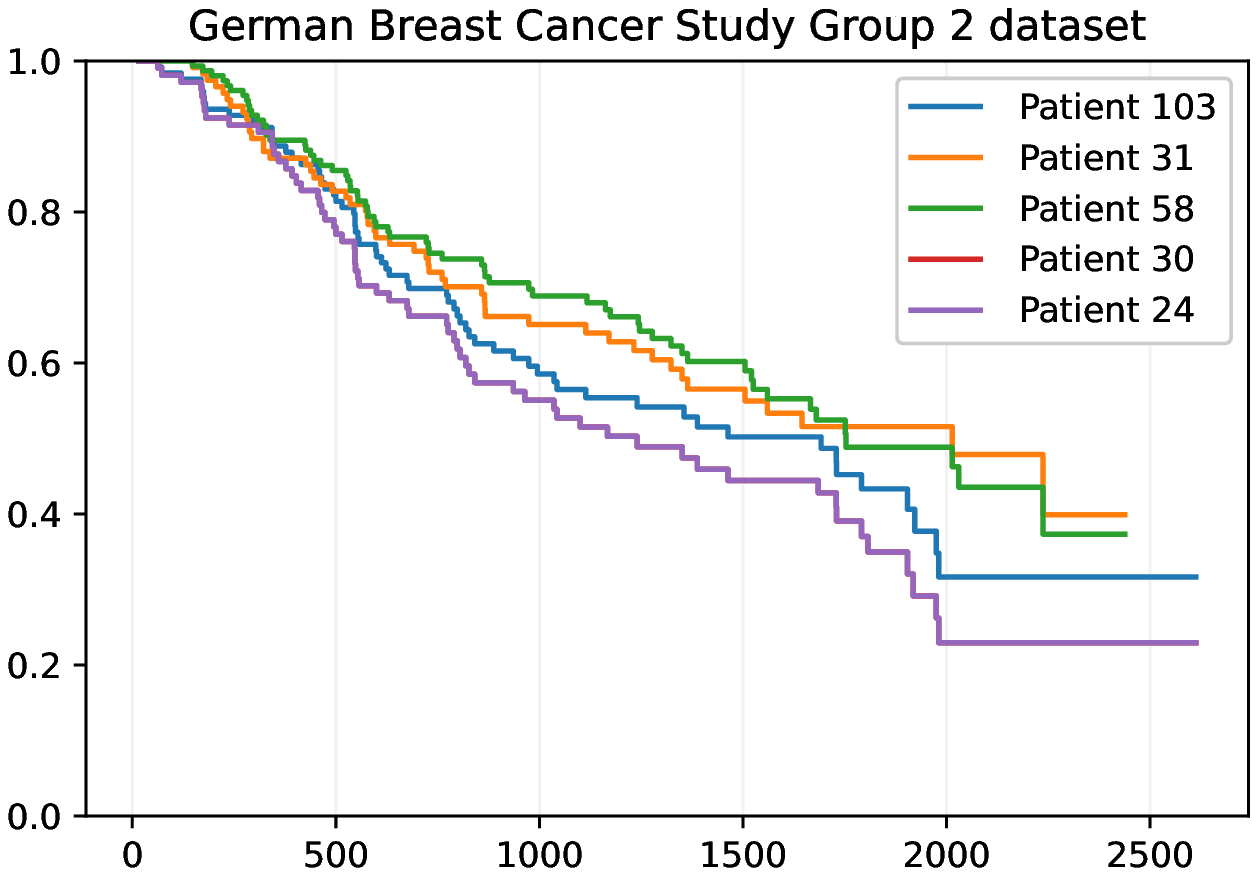}}\\
  \subfigure[$\xi_{3}$]{\includegraphics[width = 0.65\textwidth]{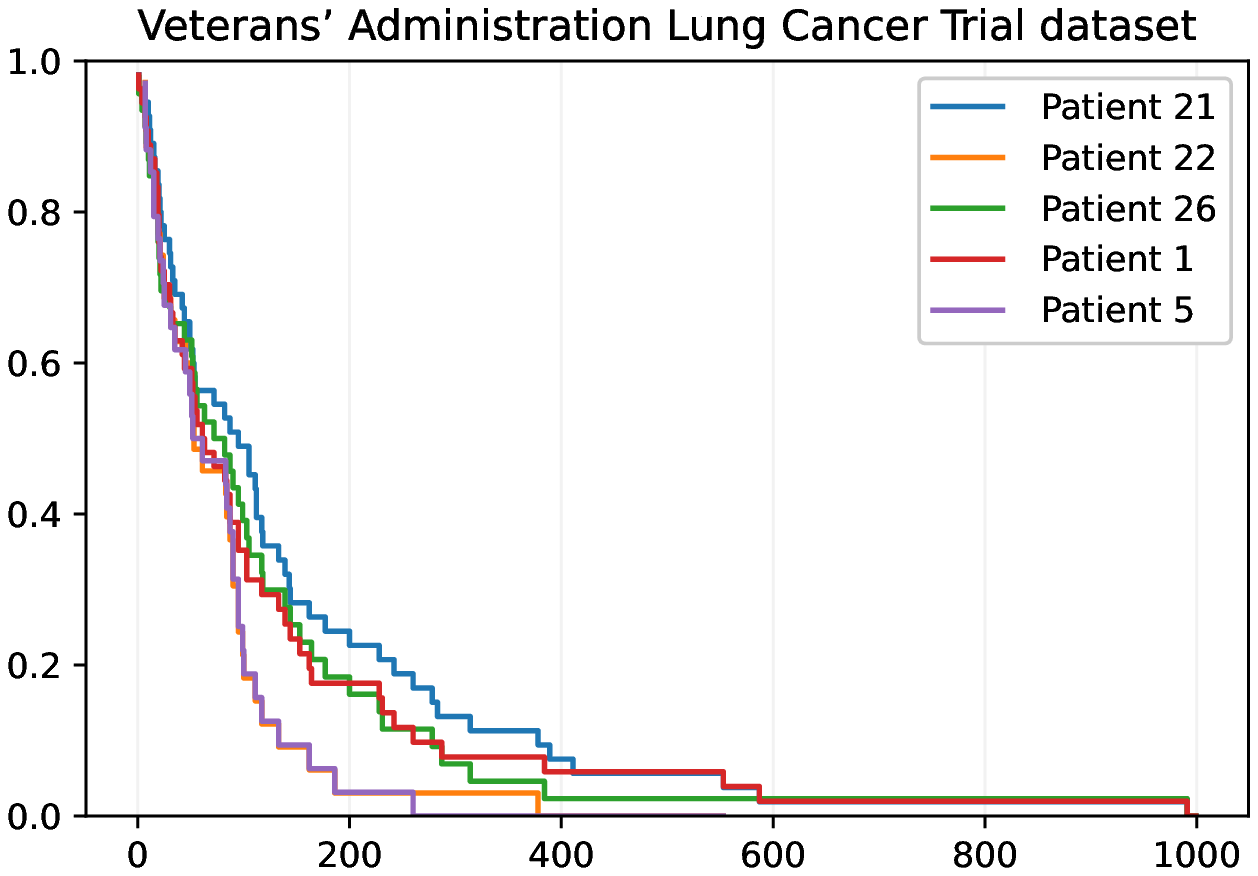}}\\
\caption{Survival Curve for Vanilla Survival Cobra for three datasets with Max\_Norm \label{fig2}}
\end{center}
\end{figure}  

\begin{figure}[ht!]
 \begin{center}
  \subfigure[$\xi_{1}$]{\includegraphics[width = 0.45\textwidth]{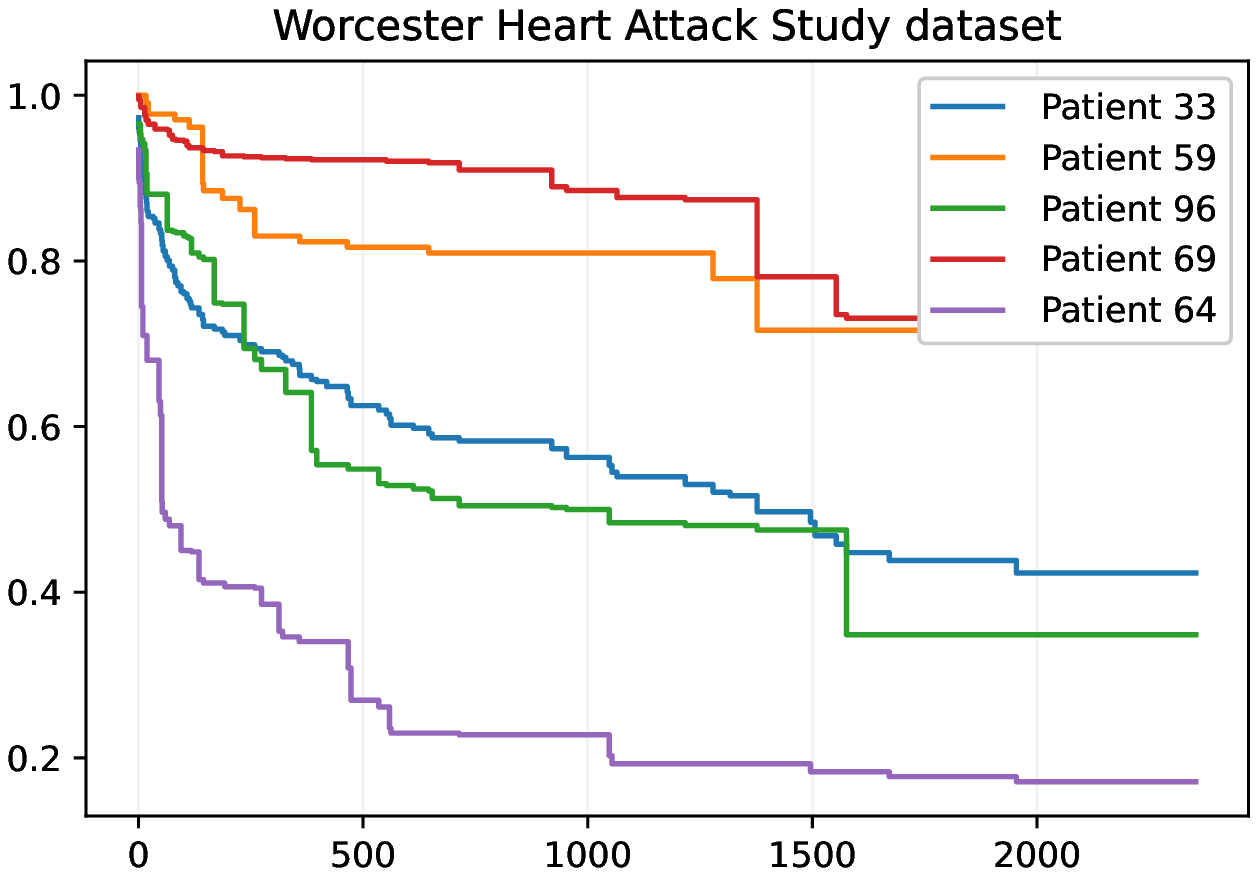}}
  \subfigure[$\xi_{2}$]{\includegraphics[width = 0.45\textwidth]{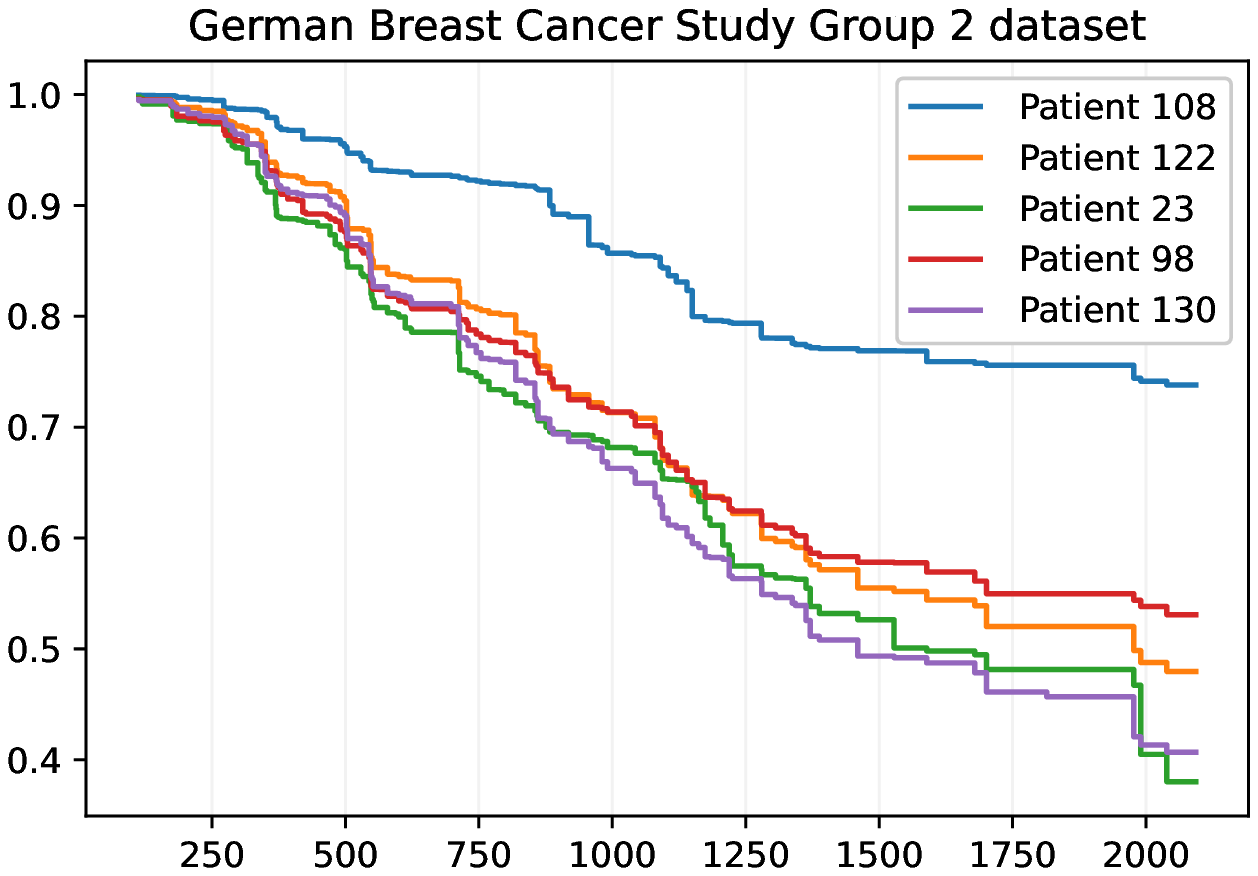}}\\
  \subfigure[$\xi_{3}$]{\includegraphics[width = 0.65\textwidth]{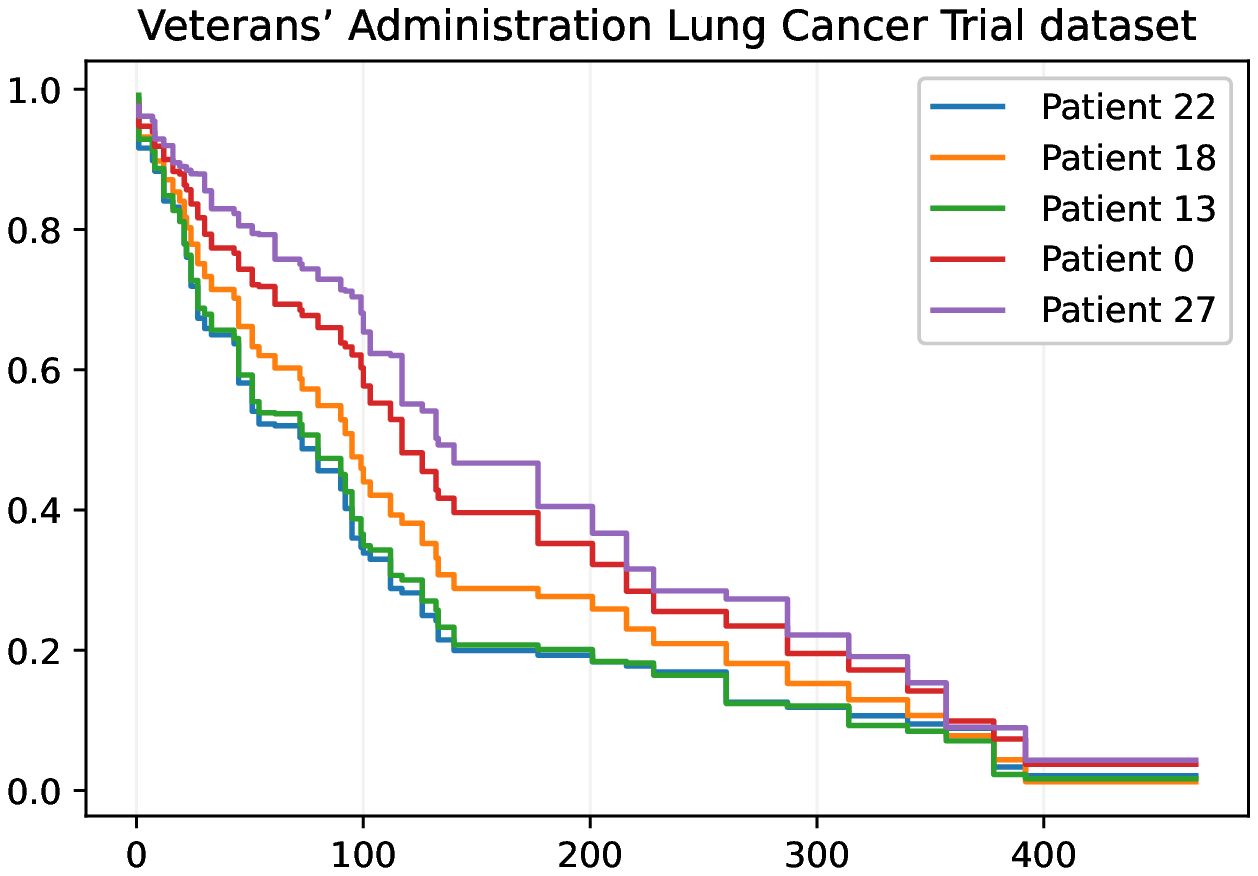}}\\
\caption{Survival Curve for Weighted Survival Cobra for three datasets with Frobenius\_Norm \label{fig3}}
\end{center}
\end{figure}

\begin{figure}[ht!]
 \begin{center}
  \subfigure[$\xi_{1}$]{\includegraphics[width = 0.45\textwidth]{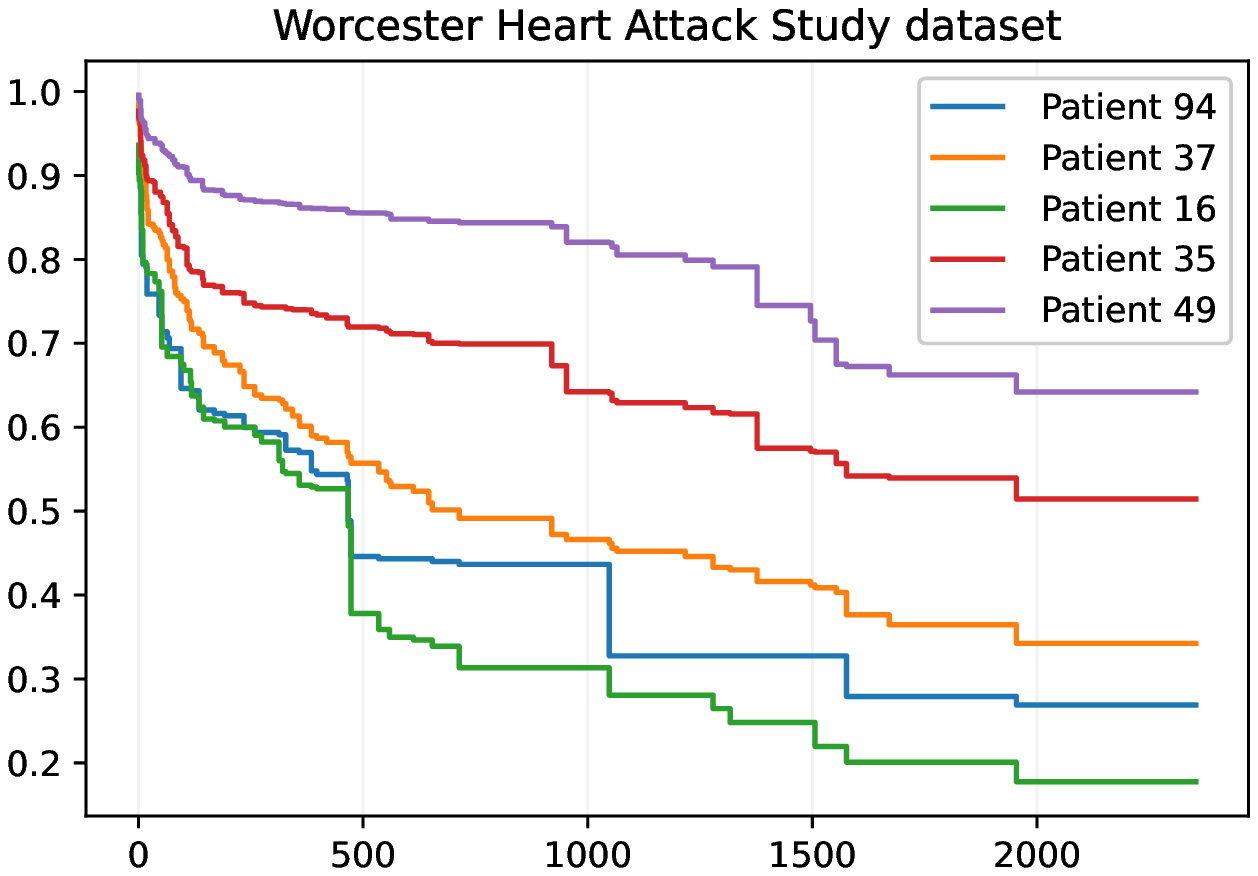}}
  \subfigure[$\xi_{2}$]{\includegraphics[width = 0.45\textwidth]{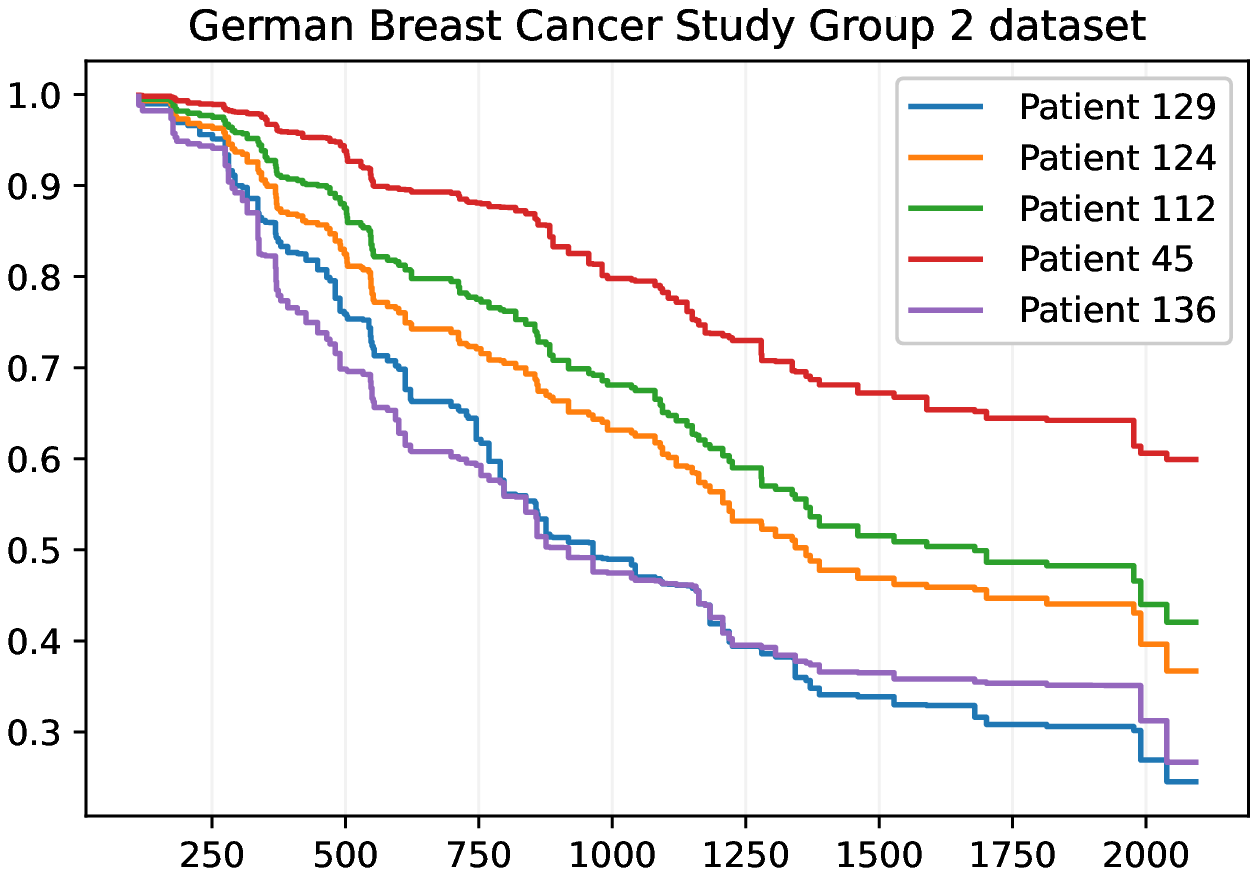}}\\
  \subfigure[$\xi_{3}$]{\includegraphics[width = 0.65\textwidth]{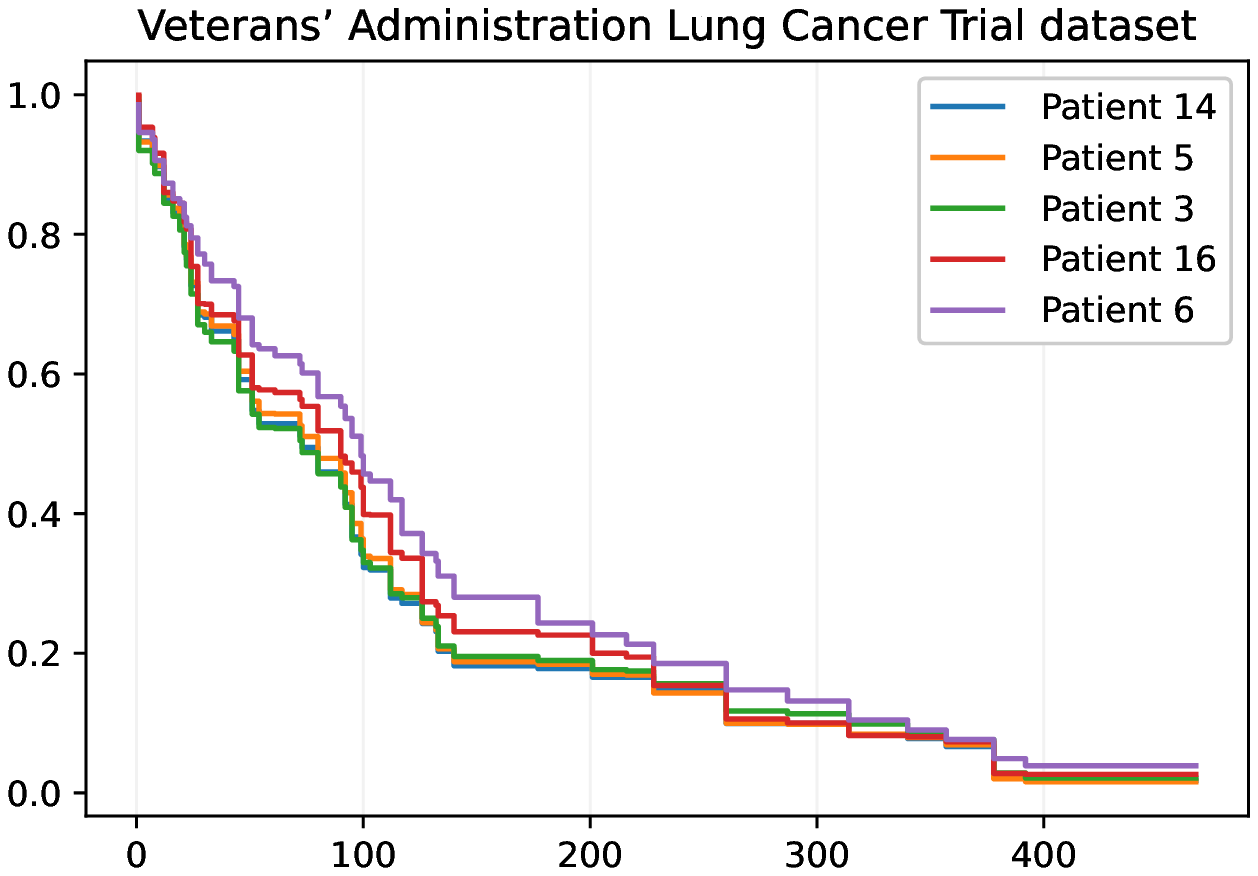}}\\
\caption{Survival Curve for Weighted Survival Cobra for three datasets with Max\_Norm \label{fig4}}
\end{center}
\end{figure}

\end{document}